\documentclass[wcp]{jmlr}

\usepackage{longtable}

\usepackage{booktabs}
\usepackage{algorithm}
\usepackage{algorithmic}
\usepackage{graphicx}
\usepackage{float}
\usepackage{multirow}
\usepackage{pgfplotstable}
\usepackage{titlesec}
\usepackage{comment}
\usepackage{amsmath,amssymb} 
\usepackage{diagbox}
\usepackage{stfloats}
\usepackage{booktabs}
\usepackage{color, soul,url}

\makeatletter
\let\Ginclude@graphics\@org@Ginclude@graphics
\makeatother

\jmlrvolume{157}
\jmlryear{2021}
\jmlrworkshop{ACML 2021}

\title[Local Aggressive Adversarial Attacks on 3D Point Cloud]{Local Aggressive Adversarial Attacks on 3D Point Cloud}

\begin{document}

\author{\Name{Yiming Sun} \Email{1220045209@njupt.edu.cn}\\
	\Name{Feng Chen} \footnotemark[1] \Email{chenfeng1271@gmail.com}\\
	\Name{Zhiyu Chen} \Email{1217053508@njupt.edu.cn}\\
	\addr Nanjing University of Posts and Telecommunications
	\AND
	\Name{Mingjie Wang} \Email{xiaowangiii@yahoo.com}\\
	\addr BNU-HKBU United International College
}
\footnotetext[1]{The first authors are Yiming Sun and Feng Chen. The corresponding author is Zhiyu Chen}
\editors{Vineeth N Balasubramanian and Ivor Tsang}

\maketitle

\begin{abstract}
	Deep neural networks are found to be prone to adversarial examples which could deliberately fool the model to make mistakes. Recently, a few of works expand this task from 2D image to 3D point cloud by using global point cloud optimization. However, the perturbations of global point are not effective for misleading the victim model.
	First, not all points are important in optimization toward misleading. Abundant points account considerable distortion budget but contribute trivially to attack.
	Second, the multi-label optimization is suboptimal for adversarial attack, since it consumes extra energy in finding multi-label victim model collapse and causes instance transformation to be dissimilar to any particular instance.
	Third, the independent adversarial and perceptibility losses, caring misclassification and dissimilarity separately, treat the updating of each point equally without a focus. Therefore, once perceptibility loss approaches its budget threshold, all points would be stock in the surface of hypersphere and attack would be locked in local optimality.
	Therefore, we propose a local aggressive adversarial attacks (L3A) to solve above issues. Technically, we select a bunch of salient points, the high-score subset of point cloud according to gradient, to perturb. Then a flow of aggressive optimization strategies are developed to reinforce the unperceptive generation of adversarial examples toward misleading victim models. Extensive experiments on PointNet, PointNet++ and DGCNN demonstrate the state-of-the-art performance of our method against existing adversarial attack methods. Our code is available at \url{https://github.com/Chenfeng1271/L3A}.
\end{abstract}
\begin{keywords}
	Point Clouds Processing, Adversarial Learning, Local Adversarial Attack
\end{keywords}


\section{Introduction}

Machine learning, especially deep learning, has achieved exceptional feats in autonomous driving, virtual reality, robotics etc. Among them, 3D sensor, such as LIDAR \cite{ref15}, which is a core ingredient of perceiving surrounding fine-grained environment, has drawn increasing attention from researchers. However, point cloud based deep learning methods \cite{ref28,ref25,ref26,ref41} are found to be vulnerable to adversarial examples where sometimes trivial movement of point cloud would mislead the deep learning model with high confidence.

Naturally, to delve the weakness of deep neural networks (DNN) \cite{ref7,ref8,ref27} toward imperceptible point cloud perturbations, adversarial attack and defense are two important topics to research. Recently, while numerous approaches have been proposed to intensively study these two tasks in 2D image domain \cite{ref10,ref11,ref12}, due to the non-rigid and sparse attribute of point cloud, this expertise may not be directly applied to the 3D counterpart. The mainstream optimization-based approaches mainly use a pair of adversarial and distortion objectives to keep the generated point cloud misleading while imperceptible \cite{ref18,ref19,ref23}. However, this kind of approaches is always ineffective because of several reasons.

\begin{figure}[t]
	\begin{center}
		\includegraphics[scale=0.54]{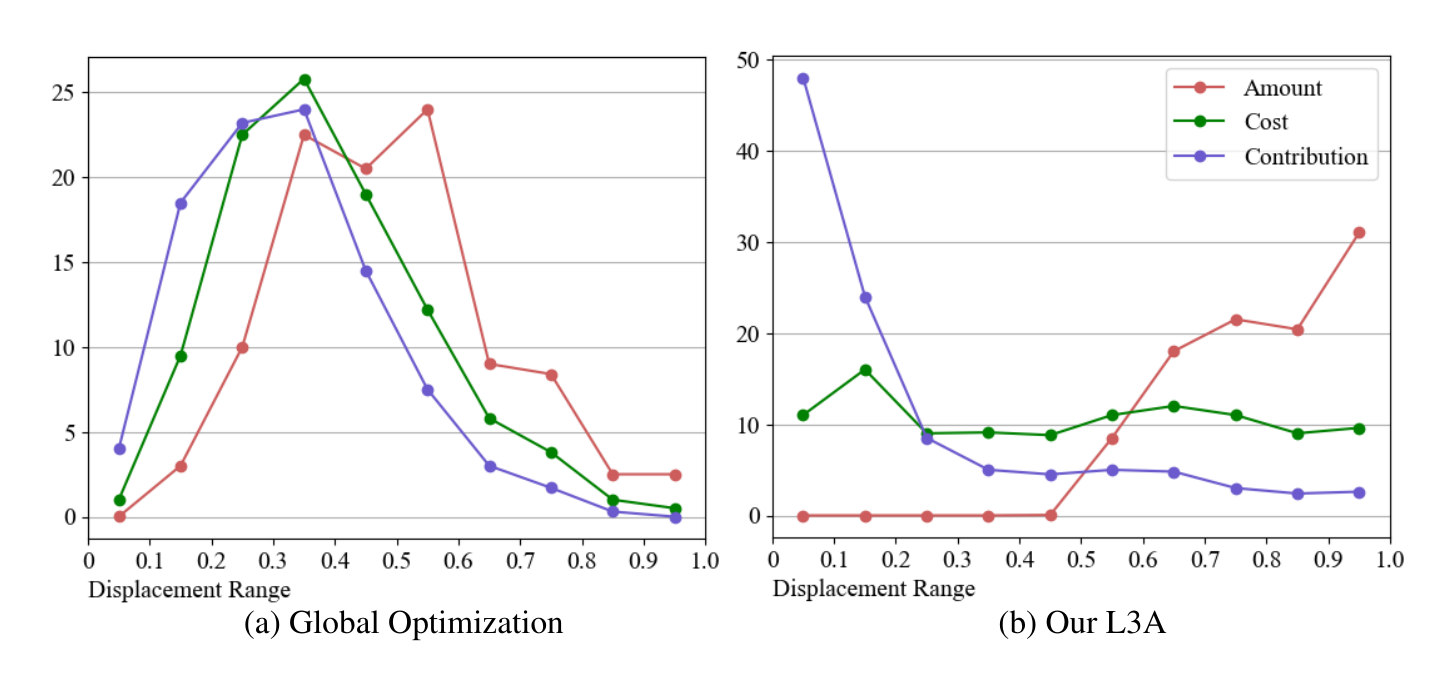}
		\caption{The statistic of amount, cost and contribution of points after attack using naive global optimization and our L3A. We divide the points in the adversarial example into 10 equal intervals according to the uniform displacement, and calculate the cost and contribution of the points. The cost of a point is defined as its displacement, and the contribution of a point is defined as the increase in classification confidence after its displacement is returned to zero. In this experiment, we set perturbation budget $\lambda$=0.3.}
		\label{fig:cc}
	\end{center}
\end{figure}
\textbf{First, not all points in the point cloud are important to adversarial attack.} Optimizing the whole points in point cloud is arduous. As shown in Figure \ref{fig:cc} (a), We divide the points into 10 intervals based on their displacement distance, and count the costs and contributions of them. In the range of 0 to 0.3, over 40\% of total points with small movement occupy considerable perceptibility budget while contribute trivially to attack. Besides, in the range of 0.6 to 1, even points move far away, they contribute only about 5\% for attack in average. This makes the tradeoff of attack and imperceptibility poor.

\textbf{Second, multi-label optimization is not the optimal choice for adversarial attack.} We observe that some examples after the attack are more likely to be identified for a particular wrong class. However, since the multi-label optimization attack always optimizes all the wrong classes, there will be competition between them. This kind of contradictory competition not only affects the attack performance, but also makes the optimization of adversarial example unstable and finally disturbs the imperceptibility. Therefore, cultivating the adversarial attack toward one preferred wrong class is more useful for attack.

\textbf{Third, adversarial loss and perceptibility loss have opposite purpose which isn't adjustable during traditional optimization.} As shown in Eq \ref{eq2}, if perceptibility loss $\mathcal{D}$ is closing to the budget $\lambda$, especially when $\lambda$ is extremely small, each point would be bounded in the surface of hypersphere and sequentially limit the attack performance. The main reason is each point equally moves toward attack purpose and greedily occupying distortion budget and none wants to withdraw.

In this paper, we propose a novel local aggressive adversarial attacks (L3A) method which is more universal and can simultaneously address these issues. The basic idea of our method is to encourage the cost-effectiveness between attack and perceptibility for each point, i.e., points moving with large displacement would contribute more for attack and vice versa.
First, we use the local salient points selection strategy. Salient points could be deemed as local salient area for classification whose transformation would be more effective for attack, instead of the whole point cloud. Then, to model the example preference as a prior knowledge, we propose a novel score loss and contractive loss to guide the adversarial examples to change towards the preferred wrong class and suppress the other weaker wrong classes in attacking process. Besides, withdrawing perturbation budget algorithm is designed to make the optimization more adjustable between attack and perceptibility to avoid it falling into local optimality.

Our key contributions can be summarized as:

1) We propose a novel local adversarial point cloud attack method named L3A which solves the problems existing in the global-optimization based attack.

2) We propose to enhance the cost-effectiveness of points between attack and perceptibility by activating the focus of optimization. By letting examples updating toward their preference and adjusting attack and indistinguishability purpose for each point, our adversarial examples could largely increase the final successful rate. 

3) Extensive experiments, including elaborate ablation study, attack and defense on ModelNet40, show that our method performs favorably against existing state-of-the-art literature on PointNet, PointNet++ and DCGNN.

\section{Related Work}
\textbf{Deep learning on point clouds.} Recently, the availability of point cloud data accelerate the development of deep learning on 3D point cloud. For projection-based methods \cite{ref38,ref39,ref40}, the dimensionality of the point cloud is compressed by projection to specific plane, such as the front view, as pseudo-image to take advantages of 2D CNN expertise, but it would inevitably cause the loss of point cloud spatial information. Compared with the projection-based method, voxel-based \cite{ref1}, \cite{ref37} and \cite{ref2} divide the 3D space into equal-spaced 3D voxels and then encode the spatial information of each voxel, but they require intensive computation which is not conducive to real-time processing. Besides, PointNet \cite{ref3} first propose to handle continuous raw point cloud directly and achieve great performance. Then PointNet++ \cite{ref4} further improves it by investing local information. Similar to PointNet++, DGCNN \cite{ref5} constructs $knn$ graphs and applies EdgeConv to capture local geometric structures. Even though these methods achieve accurate prediction of classification and recognition on large scale dataset, such as ModelNet40 \cite{ref6} and ShapeNet \cite{ref13}, they are found to be prone to adversarial attack \cite{ref15}.

\textbf{2D adversarial attack.} \cite{ref7}, \cite{ref27} and \cite{ref8} first reveal that the 2D image classification neural network could be attacked to make wrong prediction. This attack is always achieved by optimizing an objective that maximizes the prediction errors while restricting the perturbation magnitude under an $L_p$ norm. \cite{ref9}, \cite{ref10}, \cite{ref11} and \cite{ref12} are based on gradient descent to optimize the original image to achieve the goal of attack. \cite{ref22} proposes a method of generating adversarial samples under different norms $L_0$, $L_2$ and $L_{\infty}$. \cite{ref14}, \cite{ref31} and \cite{ref32} no longer disturb the whole image, but generate an adversarial patch and paste it on a specific position of the image to achieve the purpose of the attack.

\textbf{3D adversarial attack.} Recently, inspired by the 2D image attack method, many researches attempt to extend the optimization-based attack to 3D point cloud. \cite{ref34} proposes an adversarial attack to add clusters to the point cloud. \cite{ref17}, \cite{ref18} and \cite{ref19} all propose the point cloud perturbation attack approach, however, these approaches depend on updating the whole point cloud without a focus, and a plethora of points make rare contribution to attack while take considerable perturbation magnitude. In order to solve this problem, it is necessary to select appropriate salient points in the point cloud. \cite{ref20} and \cite{ref33} try to shift salient local region toward inside which is thought as point removal, instead of perturbation. In this work, we propose an effective and flexible method for local adversarial perturbation to solve this issue.

\textbf{3D adversarial defense.} As the counterpart of attack, defense technology is proposed to mitigate the harmful attack effect. $L_p$ based attack is easy to create outlying points, which could be easily detected by point removal. Classical point cloud defense methods include the algorithm based on Simple Random Sampling (SRS) \cite{ref21}, which selects several points to remove in the perturbation point cloud with medium probability. Then, Statistical Outlier Removal (SOR) \cite{ref16} is proposed by Rusu et al., which obtains better defense performance by removing outliers with $knn$. On this basis, DUP-Net \cite{ref21} deletes out-of-surface points in the perturbation point cloud, and then uses interpolation method to repair the characteristics of the point cloud, so that the point cloud can be correctly classified. Moreover, IF-Defense \cite{ref24} generates an implicit function to obtain a surface recovery module to repair the disturbed point cloud.

\section{Methodology}
Optimization-based 3D adversarial attack could be referred as distributional attack, as shown in Eq \ref{eq1}.
\begin{equation}\label{eq1}
	\underset{\hat{P}}{\operatorname{minimize}} \quad J\left(G(\theta,\hat{P}), y\right) \ \ s.t. \ \mathcal{D}(P, \hat{P})< \lambda
\end{equation}
where $P$ and $\hat{P}$ are the clean and adversarial examples, $G(\theta, P)$ is the victim DNN, $J$ is the adversarial loss and $\mathcal{D}$ is the perceptibility/dissimilar loss. $\mathcal{D}(P, \hat{P})< \lambda$ is a regulation term to avoid overshifting where $\lambda$ is perturbation budget. This equation could be reformulated as a penalty function in  {\em Lagrange} way as Eq \ref{eq2}.

\begin{equation}\label{eq2}
	\mathcal{L}_{total} = J\left(G(\theta,\hat{P}), y\right) + \kappa_1 \mathcal{D}(P, \hat{P})
\end{equation}

Besides, we divide the point set into constant subset $S_c$ and perturbation subset $S_p$ where $P\cap \hat{P} = S_c$. Namely, all the following optimization would directly work on $S_p$.

\subsection{Salient Point Selection}
In order to find salient points in the point cloud, we take the gradient of cross entropy loss $\mathcal{L}_{ce}$ in the victim model $G$ as the instance-wise importance of points. According to gradient magnitude $\frac{\partial \mathcal{L}_{ce}}{\partial p{i}}$ where $p_i \in P$, we select the top $m$ vertexes and use group sampling ($knn$) of PointNet to further locate local point clusters where each cluster has $n$ points in theory.

\subsection{Score Loss and Contractive Loss}
The primary goal of the point cloud optimization-based attack is making the victim model misclassify the examples. Inspired by the 2D adversarial methods, we first propose a basic loss to directly reduce the confidence of the victim model on the true label of the examples. As shown in Eq \ref{eq3}, we choose to use normalized estimation probability and ground truth as straightforward attack loss, where $k$ enumerates the total classes, $y$ is the ground truth for one-hot encoding and $\hat{y}$ is the prediction probability.

\begin{equation}\label{eq3}
	\mathcal{L}_{base} = \sum_{k}{y_k\hat{y}_k}\\
\end{equation}

Moreover, the final appearance and classification result of adversarial examples largely depends on instance-level preference, e.g., some examples of airplane may have a preference for being cars, but not all airplane instances have this preference. We rethink this instance-level preference as a prior knowledge and delve it in optimization.
In multi-label optimization, the preferred wrong class would be engaged in competing with other non-preferred wrong classes, which may be detrimental to convergence and sequentially affect the final performance.
Therefore, in Eq. \ref{eq4}, we introduce single-label optimization loss $\mathcal{L}_{score}$ where the wrong class with highest confidence is selected as the explicit direction in overall optimization.
Moreover, as shown in Eq. \ref{eq5}, to release the preferred wrong class from competition, we propose intra-class contractive loss $\mathcal{L}_{cons}$ that not only encourages the confidence of preferred wrong class $\hat{y}_{pre}$ to be larger, but also compresses the wealth of confidence occupied by non-preferred wrong classes $\hat{y}_{non-pre}$.

\begin{equation}\label{eq4}
	\mathcal{L}_{score} = 1 - \max_{k}{(1-y_k) \hat{y}_k}
\end{equation}

\begin{equation}\label{eq5}
	\begin{split}
		\mathcal{L}_{cons}= &\sum_k{y_k\cdot \max\{\lambda -\hat{y}_{pre}+\hat{y_k}, 0\}} \\
		&+1-\frac{\hat{y}_{pre}}{\sum{\hat{y}_{non-pre}}}
	\end{split}
\end{equation}

In early optimization, our score loss and contractive loss would model the instance preference in a static way and then quickly find and stabilize the optimization direction.
Overall, the total attack loss $J$ is composed by $\mathcal{L}_{base}$, $\mathcal{L}_{score}$ and $\mathcal{L}_{cons}$, as shown in Eq \ref{eq6} where $\kappa_2$ and $\kappa_3$ are the balance factors and set to 1 and 0.5 respectively.

\begin{equation}\label{eq6}
	J = \mathcal{L}_{base} + \kappa_2 \mathcal{L}_{score} + \kappa_3 \mathcal{L}_{cons}
\end{equation}

\subsection{Perceptibility Loss}
There are several ways to measure the distance (perceptibility) of two point sets. We adopt three kinds of perceptibility loss, i.e., $L_2$ distance, Chamfer distance \cite{ref30} and Hausdorff \cite{ref29} distance, as perceptibility loss $\mathcal{D}$. Given two point sets $A$ and $B$, $\mathcal{D}_{chamfer}$, $\mathcal{D}_{Hausdorff}$ and $\mathcal{D}_{L_2}$ could be represented as:
\begin{equation}\label{eq7}
	\begin{split}
		\mathcal{D}_{chamfer}(A, B) &=\frac{1}{|A|} \sum_{a \in A} \min _{b \in B}\|a-b\|_{2} \\
		&+\frac{1}{|B|} \sum_{b \in B} \min _{a \in A}\|a-b\|_{2}
	\end{split}
\end{equation}
\begin{equation}\label{eq8}
	\mathcal{D}_{Hausdorff}(A, B)=\max_{a\in A} \ \min_{b\in B} ||a-b||_2
\end{equation}

\begin{equation}\label{eq9}
	\mathcal{D}_{L_2} = \sum_{i=1}^{|A|}{||A_i - B_i||_2}
\end{equation}

In our L3A, $\mathcal{D}_{L_2}$ only calculate pair-wise points in $S_p$ ($|S_p|<<|P|$) for faster training speed. However, we find that this distance may lead to outliers due to the lack of information about the surrounding points. Chamfer and Hausdorff distances enumerate the whole point set to find the closest points which consumes $\mathcal{O}( |P|\times |S_p|)$. The difference is that the chamfer distance takes the average for all the nearest pairs but the Hausdorff distance takes the maximum for them. Instead of calculating the distance between the point pairs directly, Chamfer and Hausdorff distances are based on two complete point sets, thus the original shape of the point cloud can be retained as much as possible.

\subsection{Perturbation-budget Withdrawing Algorithm}
It's worth noting that the attack loss in initial few iterations is extremely large  which impetuously push the points to shift. However, after a while, the perturbation budget is almost consumed and following shift of points is limited on the surface of hypersphere, as shown in Eq. \ref{eq1}. Therefore, we aim to force some points which contribute trivially in current iteration to move backward to its anchor position $x_i^{\prime}$ to release some free budget for the moving of important points. With respect to $\mathcal{D}$, the anchor position of a point is remembered at latest distance calculation. For example, as shown in Eq. \ref{eq9}, $\mathcal{D}_{L_2}$ uses corresponding points in clean and adversarial examples, the anchor position of point $a_i$ is consistently to be $b_i$. While in $\mathcal{D}_{chamfer}$, each anchor location in clean example is calculated by its equation. As shown in the Perturbation-budget Withdrawing Algorithm (PWA) of Algorithm \ref{alg:algorithm}, we first push trivial points back with a minor step and then erase the gradient of them to avoid generating outlying points which would be easily removed by outlier defense. The next gradient-erasing step would provide more freedom in current iteration. Furthermore, the released budget by the moving-back step would work in next iteration. Essentially, this strategy is greedy and we will discuss the tradeoff between greedy attack and imperceptibility in Section \ref{Sec:pwa}.

\begin{algorithm}[t]
	\caption{Perturbation-budget Withdrawing Algorithm (PWA)}
	\label{alg:algorithm}
	\textbf{Input}: Attack loss $J$,  perceptibility loss $\mathcal{D}$, perturbation subset $S_p$.\\
	\textbf{Parameter}: Total iteration number $N$, activation threshold $h$ to use PWA, selection threshold $w$ for filtering unimportant points, perceptibility budget $\lambda$, spatial position $x_i\in \mathbb{R}^3$ of $p_i\in S_p$, anchor position $x_i^{\prime}$ of $p_i$, learning rate $lr$.\\
	\textbf{Output}: Regulated gradient $\frac{\partial J}{\partial p_{i}}$ of points $p_i$ and adjusted point coordinate $x_i$.
	\begin{algorithmic}[1] 
		\WHILE{0.2$N<iteration<N$}
		\IF {$\mathcal{D} > \lambda - h$}
		\STATE{record $x_i^{\prime}$}
		\IF {$|\frac{\partial J}{\partial p_{i}}|< w$}
		\STATE $x_i = x_i - (x_i - x_i^\prime) \cdot lr \cdot \gamma |\frac{\partial J}{\partial p_{i}}|$
		\STATE $\frac{\partial \mathcal{L}_{total}}{\partial p_{i}}$ = 0.
		\ELSE
		\STATE pass
		\ENDIF
		\STATE $iteration= iteration + 1$
		\ELSE
		\STATE $iteration= iteration + 1$
		\ENDIF
		\ENDWHILE
		\STATE \textbf{return}  gradient $\frac{\partial J}{\partial p}$, spatial position $x$
	\end{algorithmic}
\end{algorithm}

\section{Experimental Results}

\subsection{Experimental Setup}
\textbf{Dataset and victim model.} We use the ModelNet40 \cite{modelnet40} to evaluate the performance of our method, including training, testing the victim models and generating the adversarial examples. This dataset consists of 12,311 CAD models with 40 common object categories. These CAD models are divided into 9,843 for training and 2,468 for testing. The test split is used for evaluating the attack performance. Besides, we use PointNet \cite{ref3}, PointNet++ \cite{ref4} and DGCNN \cite{ref5} as our victim models which are trained in official code with default hyperparameters.

\textbf{Implementation Details.}  The point cloud contains 10,000 points, and we sample 1,024 points for training. The total epochs is set to 200. We use the Adam optimizer to optimize the objectives where the learning rate $lr$ is set to 0.001, and the momentum $\beta_{1}$ and $\beta_{2}$ are set to 0.9 and 0.999 respectively. We warm up the initial optimization within the first $n$ epochs (about 0.2 times of the total epochs). The rest epochs would follow the proposed strategy. Warm-up aims to stabilize the instance-level preference, and specifically find the preferred class.

\begin{table}[t]
	\footnotesize{
		\begin{center}
			\caption{Attack success rate of L3A using different perturbation loss on PointNet, PointNet++ and DGCNN.}
			\label{fig:zong}
			\begin{tabular}{c|c|ccc|ccc|ccc}
				\hline
				\multicolumn{1}{r|}{\multirow{2}{*}{Victim Model}} & \multirow{2}{*}{$\lambda$ } & \multicolumn{3}{c|}{$\mathcal{D}_{L_2}$} & \multicolumn{3}{c|}{$\mathcal{D}_{Chamfer}$} & \multicolumn{3}{c}{$\mathcal{D}_{Hausdorff}$}                                                                                                                                                                \\
				\multicolumn{1}{r|}{}                              &                             & \multicolumn{1}{c}{None}                 & \multicolumn{1}{c}{SOR}                      & \multicolumn{1}{c|}{SRS}                      & \multicolumn{1}{c}{None} & \multicolumn{1}{c}{SOR} & \multicolumn{1}{c|}{SRS} & \multicolumn{1}{c}{None} & \multicolumn{1}{c}{SOR} & \multicolumn{1}{c}{SRS} \\ \hline
				\multirow{3}{*}{PointNet}                          & 0.001                       & 99.7\%                                   & 92.6\%                                       & 99.6\%                                        & 99.7\%                   & 92.7\%                  & 99.7\%                   & 99.7\%                   & 91.5\%                  & 99.7\%                  \\
				                                                   & 0.0025                      & 99.7\%                                   & 92.7\%                                       & 99.6\%                                        & 99.7\%                   & 92.7\%                  & 99.7\%                   & 99.7\%                   & 92.4\%                  & 99.7\%                  \\
				                                                   & 0.005                       & 99.7\%                                   & 92.7\%                                       & 99.6\%                                        & 99.7\%                   & 92.7\%                  & 99.7\%                   & 99.7\%                   & 92.4\%                  & 99.7\%                  \\ \hline
				\multirow{3}{*}{PointNet++}                        & 0.001                       & 97.0\%                                   & 94.0\%                                       & 88.5\%                                        & 97.2\%                   & 94.0\%                  & 89.2\%                   & 97.0\%                   & 92.9\%                  & 90.0\%                  \\
				                                                   & 0.0025                      & 96.9\%                                   & 93.2\%                                       & 90.5\%                                        & 97.3\%                   & 92.4\%                  & 89.2\%                   & 97.4\%                   & 94.4\%                  & 89.4\%                  \\
				                                                   & 0.005                       & 96.8\%                                   & 93.3\%                                       & 89.5\%                                        & 97.0\%                   & 93.4\%                  & 89.7\%                   & 97.3\%                   & 92.9\%                  & 89.3\%                  \\ \hline
				\multirow{3}{*}{DGCNN}                             & 0.001                       & 98.8\%                                   & 97.0\%                                       & 97.0\%                                        & 99.0\%                   & 96.0\%                  & 97.4\%                   & 98.9\%                   & 95.7\%                  & 97.5\%                  \\
				                                                   & 0.0025                      & 98.9\%                                   & 95.8\%                                       & 95.8\%                                        & 99.0\%                   & 96.0\%                  & 97.1\%                   & 98.9\%                   & 96.0\%                  & 97.6\%                  \\
				                                                   & 0.005                       & 99.0\%                                   & 96.0\%                                       & 96.0\%                                        & 99.0\%                   & 96.0\%                  & 97.7\%                   & 98.9\%                   & 96.4\%                  & 97.8\%                  \\ \hline
			\end{tabular}
		\end{center}}
\end{table}

\subsection{Attack and Defense}
We report the performance of our method with various settings on PointNet \cite{ref3}, PointNet++ \cite{ref4} and DGCNN \cite{ref5} in Table \ref{fig:zong}. Our method could achieve promising results on all three victim models under no defense, SOR and SRS: L3A obtains 99.7\%/97.1\%/98.9\% success rate on three model respectively without using defense strategy in average. Besides, after the defense of SOR/SRS, our model only decreases 7.2\%/0.1\%, 3.7\%/4.8\% and 2.8\%/1.8 on three models averagely and all results are all over 92\%, indicating our L3A is robust to defense. Moreover, we adjust the magnitude of $\lambda$ from 0.001 to 0.005 which is extremely small, but the performance only has minor fluctuation.

Moreover, Figure \ref{fig:cc} illustrates the cost-effectiveness comparison with and without using our method (i.e., L3A and global optimization with $\mathcal{D}_{L_2}$). Intuitively, we believe points moving far away from anchor and occupying large budget would contribute more to attack. However, in Figure \ref{fig:cc} (a), over 46\% percent of total points concentrate on 0.1-0.3 range where each point moves trivially, but they cost about 33\% budget and only contribute 14\% for attack. Besides, in the displacement range of 0.6 to 1.0 where per-point displacement is rigid, their contribution also decline to 3\%, which is contrary to our intuition. Namely, this Gaussian-like distributions of three lines mean global optimization doesn't allocate reasonable budget to each point from low to high intervals. In Figure \ref{fig:cc} (b), we also evaluate the cost-effectiveness of each points of our method. It shows that our method inhibits the cost of low range (0-0.3) and push over 78\% points to locate in this range, which can avoid appreciable point transformation and keep original appearance as much as possible. Meanwhile, in the high range (0.6-1.0), the contribution of points in this range largely increases with stable cost, indicating the each further step the point made would effectively lead to mislead the victim model.

\begin{figure}[!t]
	\begin{center}
		\includegraphics[scale=0.5]{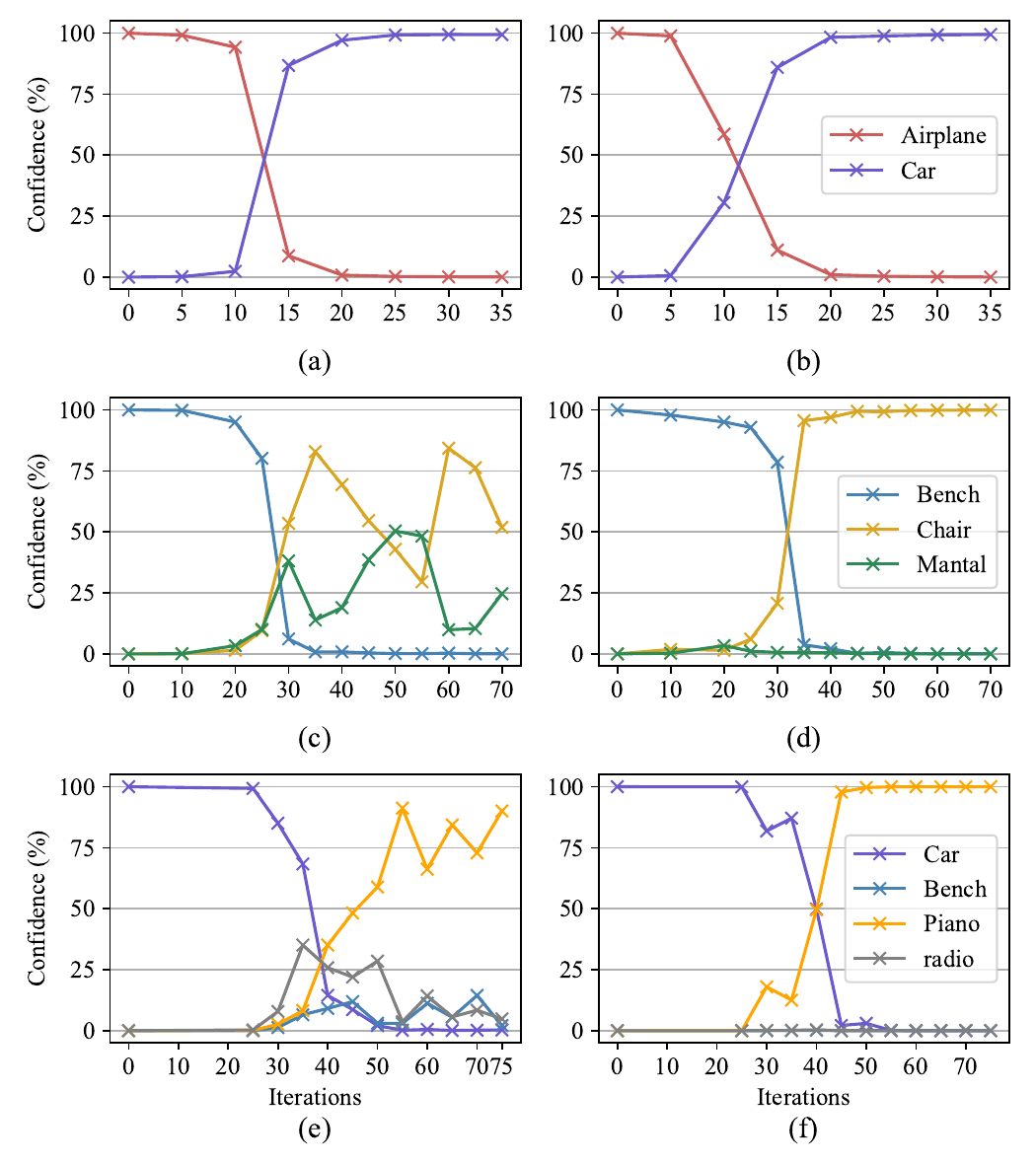}
		\caption{A comparison of optimization process of three examples (airplane, bench, car in three rows separately) with and without using $\mathcal{L}_{score}$ and $\mathcal{L}_{cons}$. Subfigure (a), (c), (e) show three examples without using our method, the airplane converges to the best class, while the other two examples have poor convergence. Using our method, all examples could converge quickly to the best class as shown in Subfigure (b), (d), (f).}
		\label{fig:sla}
	\end{center}
\end{figure}

\subsection{Single-label Attack and Withdrawing Algorithm}\label{Sec:pwa}
As shown in Figure \ref{fig:sla}, we report the convergence of optimizing airplane, bench, car toward adversarial examples by only using $\mathcal{L}_{base}$ (left column) and using our $\mathcal{L}_{score}$ and $\mathcal{L}_{cons}$ (right column). Notably, our method could not only find the class-level updating direction toward specific class in Figure \ref{fig:sla} (a \& b), but also suppress all potential weak wrong classes when it has two strong competitive wrong classes or is harassed by considering multiple class choice. Moreover, in the pair of Figure \ref{fig:sla} (c \& d) and (e \& f), our method could faster the convergence of all examples and achieve higher confidence for preferred class.

\begin{figure}[!h]
	\begin{center}
		\includegraphics[scale=0.5]{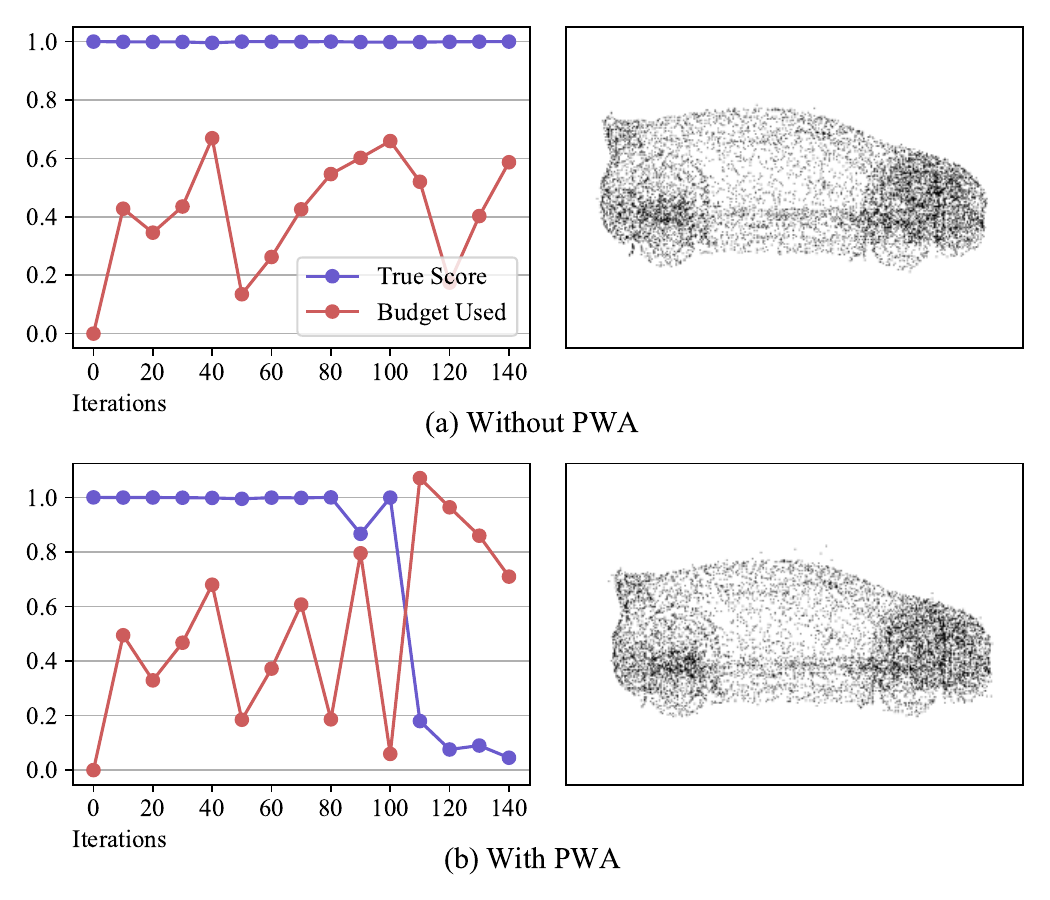}
		\caption{Attack results on car with and without using our PWA.} 
		\label{fig:withdraw}
	\end{center}
\end{figure}

\begin{figure}[!t]
	\begin{center}
		\includegraphics[scale=0.23]{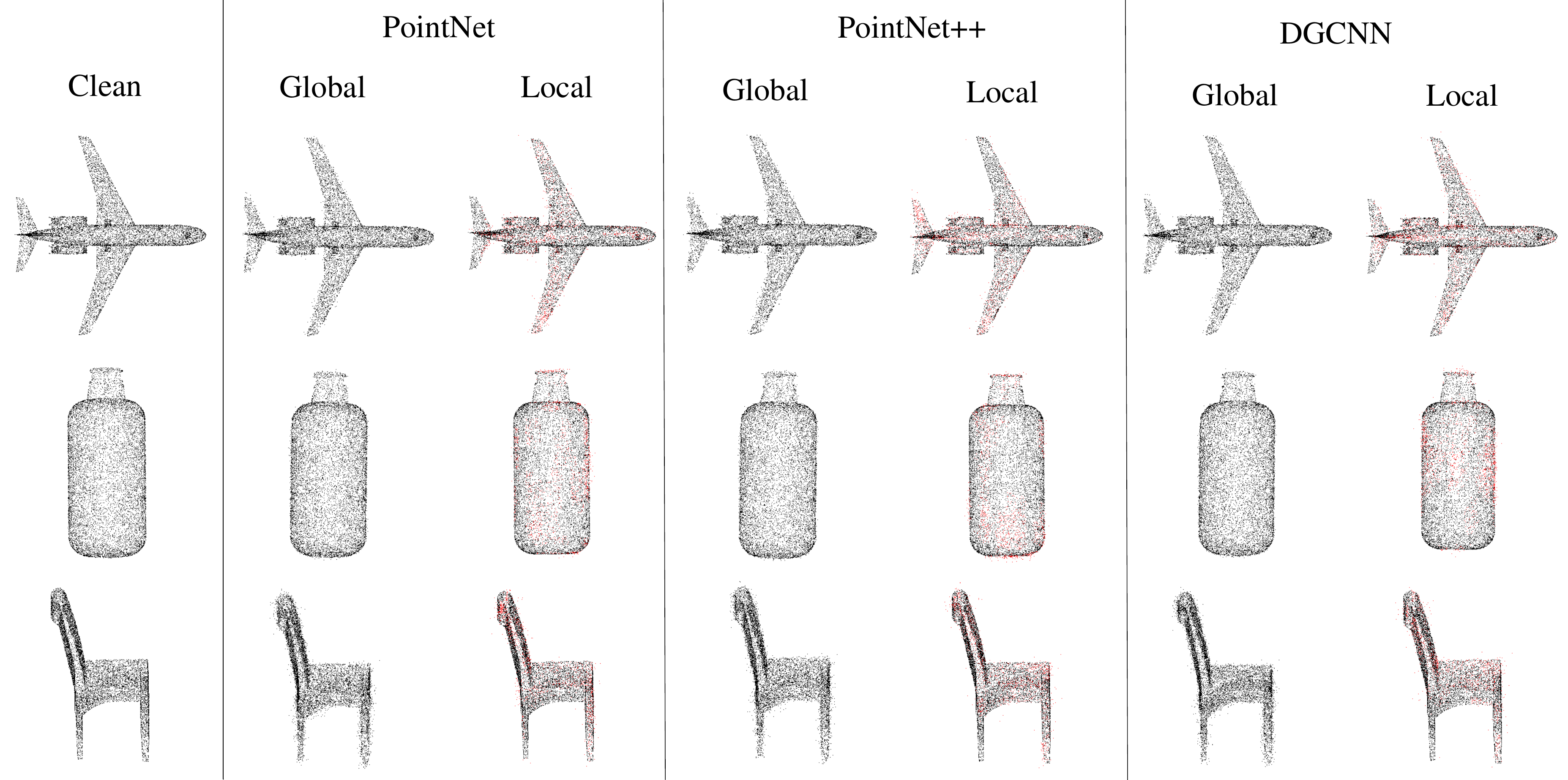}
		\caption{Visualization of adversarial examples of PointNet, PointNet++, and DGCNN models which are attacked by global optimization and our L3A. The points selected in the local point selection are marked red in the figure.}
		\label{fig:visu}
	\end{center}
\end{figure}
\begin{figure}[!h]
	\begin{center}
		\includegraphics[scale=0.07]{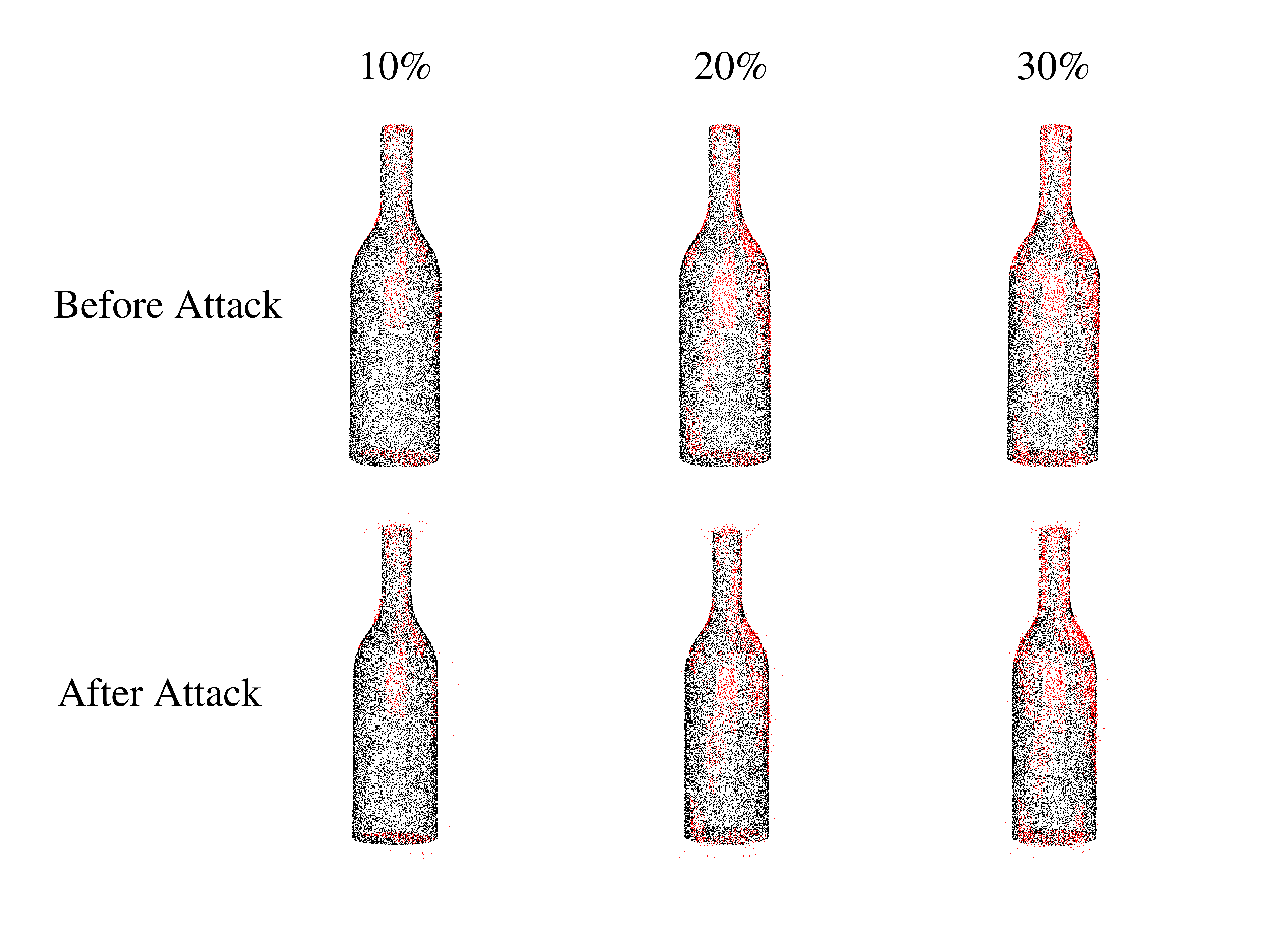}
		\caption{Visualization of different proportions of selected points. Salient points marked in red are selected at 10\%, 20\%, and 30\% ratios in PointNet model.} 
		\label{fig:count_visu}
	\end{center}
\end{figure}

\begin{table}[h]
	\begin{center}
		\caption{Ablation study of our L3A on PointNet, PointNet++ and DGCNN using score loss, contractive loss, PWA and local salient point selection.}
		\label{tab:ablation}
		\setlength{\tabcolsep}{2mm}{
			\begin{tabular}{c|cccc|ccc}
				\hline
				\multirow{2}{*}{Group} & \multicolumn{4}{c|}{Method} & \multicolumn{3}{c}{Victim Model}                                                    \\
				                       & $\mathcal{L}_{score}$       & $\mathcal{L}_{cons}$             & PWA      & Local    & PN      & PN++   & DGCNN   \\ \hline
				1                      & $\times$                    & $\times$                         & $\times$ & $\times$ & 98.3\%  & 96.1\% & 100.0\% \\
				2                      & $\surd$                     &                                  &          &          & 98.8\%  & 97.2\% & 100.0\% \\
				3                      &                             & $\surd$                          &          &          & 100.0\% & 99.9\% & 100.0\% \\
				4                      &                             &                                  & $\surd$  &          & 98.3\%  & 96.3\% & 100.0\% \\
				5                      &                             &                                  &          & $\surd$  & 97.2\%  & 94.0\% & 98.0\%  \\
				6                      & $\surd$                     & $\surd$                          &          &          & 100.0\% & 99.9\% & 100.0\% \\
				7                      &                             & $\surd$                          & $\surd$  &          & 100.0\% & 99.9\% & 100.0\% \\
				8                      &                             &                                  & $\surd$  & $\surd$  & 97.2\%  & 94.5\% & 98.2\%  \\
				9                      & $\surd$                     & $\surd$                          & $\surd$  &          & 100\%   & 100\%  & 100\%   \\
				10                     &                             & $\surd$                          & $\surd$  & $\surd$  & 99.3\%  & 99.9\% & 99.3\%  \\
				11                     & $\surd$                     &                                  & $\surd$  & $\surd$  & 98.2\%  & 96.7\% & 99.4\%  \\
				12                     & $\surd$                     & $\surd$                          &          & $\surd$  & 99.6\%  & 97.1\% & 98.9\%  \\
				13                     & $\surd$                     & $\surd$                          & $\surd$  & $\surd$  & 99.3\%  & 98.9\% & 99.9\%  \\
				\hline
			\end{tabular}}
	\end{center}
\end{table}

\begin{table}[htb]
	\begin{center}
		\caption{The effect of point ratio on performance. $m$ represents the number of selected vertexes, and $n$ represents the number of point of cluster around the vertexes using the $knn$.}
		\label{tab:spr}
		\setlength{\tabcolsep}{1.5mm}{
			\begin{tabular}{cc|cccc}
				\hline
				$m$ & $n$ & Proportion & PointNet & PointNet++ & DGCNN  \\ \hline
				30  & 30  & 25.06\%    & 95.5\%   & 89.9\%     & 95.2\% \\
				30  & 40  & 25.92\%    & 95.5\%   & 90.5\%     & 95.1\% \\
				30  & 50  & 26.56\%    & 95.5\%   & 90.1\%     & 94.9\% \\
				40  & 30  & 32.17\%    & 96.2\%   & 91.7\%     & 97.3\% \\
				40  & 40  & 32.28\%    & 96.2\%   & 92.0\%     & 97.4\% \\
				40  & 50  & 33.53\%    & 96.3\%   & 91.9\%     & 97.0\% \\
				50  & 30  & 37.51\%    & 96.8\%   & 92.8\%     & 98.4\% \\
				50  & 40  & 39.22\%    & 96.8\%   & 92.6\%     & 98.4\% \\
				50  & 50  & 39.53\%    & 96.8\%   & 92.9\%     & 98.5\% \\ \hline
			\end{tabular}}
	\end{center}
\end{table}

In addition, the adversarial loss of some examples may not further decrease, because these examples are regulated by perceptibility loss $\mathcal{D}$ and trapped in local optimality. As shown in Figure \ref{fig:withdraw} (a), the budget is consumed in early stage. At 50 iteration, the budget is released due to self-adjustment, but following optimization still fail to make the attack succeeded. Our PWA solves this issue by encouraging a part of unimportant points to release budget to escape this local optimality. Figure \ref{fig:withdraw} (b) shows that at 90 iteration, our PWA force points to release about 0.7 budget in a sudden, and pending the confidence could fall down to 0.08.
Although our PWA algorithm is a greedy strategy, it does not heavily disturb perceptibility, since the PWA doesn't trigger continuously but is always in the withdraw-adjustment cycle. This gives all the points a chance to be adjusted to a better position in the next iteration. As shown in the right part of Figure \ref{fig:withdraw}, even after heavy budget withdrawing, the appearance of point cloud is still reasonable like a car.

\subsection{Ablation Study}
We evaluate each component of our method, and the results are shown in the Table \ref{tab:ablation}. Compared the group 1 to 4, the proposed $\mathcal{L}_{score}$, $\mathcal{L}_{cons}$ and PWA could bring 0.5\%/1.1\%, 1.7\%/3.8\% and 0/0.3\% improvements on PointNet/PointNet++ respectively in global optimization case. However, when only using the local salient point selection, i.e., in group 5, the performance meets 1.1\%/2.1\%/2.0\% decrease on three models. This means only using local points to optimization may not a good choice. In addition, we further assess $\mathcal{L}_{score}$, $\mathcal{L}_{cons}$ and PWA in local optimization case. As shown in group 5 and 8, solely using PWA would bring 0.5\%/0.2\% gains on PointNet++ and DGCNN, and similarly, the fully armed setting could achieve further improvement. It is worthy to note that the best performance of this table uses variants with global optimization settings, i.e., group 6 and 7, instead of our fully armed setting. However, our method is only 0.7\%/1.0\%/0.1\% lower than them and most points don't move after attack, which is more similar to original examples in visual appearance. We also visualize the point cloud in Figure \ref{fig:visu} where selected points using our method are marked in red. Our L3A keeps most points motionless and drives the pending points to move surrounding the original outlook. Therefore, our method could largely attack the victim model while transforming the point cloud indistinguishably.

\begin{table}[!t]
	\begin{center}
		\caption{Comparison between our method and some global and local attack methods on PointNet, PointNet++ and DGCNN.}
		\label{tab:compare}
		\small{
			\begin{tabular}{c|c|c|c|c|c|c|c|c|c}
				\hline
				\multirow{2}{*}{} & ours         & \multicolumn{5}{c|}{global} & \multicolumn{3}{c}{local}                                                       \\ \cline{2-10}
				                  & L3A          & RP                          & GP                        & Chamfer & EoTPG
				                  & Xiang et al. & Zheng et al.
				                  & Kim et al.
				                  & Nudge                                                                                                                        \\
				\hline
				Pointnet          & 99.3\%       & 12.4\%                      & 89.8\%                    & 100.0\% & 40.5\% & 100\% & 55.7\% & 89.4\% & 97.4\% \\ \hline
				Pointnet++        & 98.9\%       & 9.4\%                       & 98.8\%                    & 97.7\%  & 87.8\% & -     & 41.5\% & 88.8\% & -      \\ \hline
				DGCNN             & 99.9\%       & 11.6\%                      & 34.5\%                    & 99.6\%  & 23.3\% & -     & 35.8\% & 62.2\% & 69.5\% \\ \hline
			\end{tabular}}
	\end{center}
\end{table}
We also elaborate on the influence of selection point ratios to the ultimate results. In Table \ref{tab:spr}, we select different combinations of cluster and sample number $m$ and $n$. It is obvious that with the increase of proportion, the performance of our method on three victim models accordingly rise. Moreover, even using similar but reverse setting, such as $n=30$ \& $m=40$ and $n=40$ \& $m=30$, the proportion of them is different, since most cluster would concentrate on salient region and therefore the overlapped points would increase with cluster size. As shown in Figure \ref{fig:count_visu}, we visualize the selected points in bottle. Most of selected points locate in the mouth, upper part and bottom where are the salient region.

\subsection{Comparison of Other Methods}
As shown in Table \ref{tab:compare}, we first compare the performance of our L3A with existing global-optimization-based attack methods, including random perturbation (RP), gradient projection attack (GP) \cite{ref23}, Chamfer attack \cite{ref23}, EoTPG \cite{ref19} and Xiang et al. \cite{ref34} on PointNet, PointNet++ and DGCNN, and then compare our method with some methods based on local optimization including Zheng et al. \cite{ref20}, Kim et al. \cite{ref42} and Nudge \cite{ref43}. In the global optimization method, our L3A achieves the best performance on PointNet++ and DGCNN and is just 0.7\% lower than Chamfer and Xiang et al. on PointNet. Moreover, it outperforms Chamfer and EoTPG with 1.2\% and 11.1\% gain on PointNet++ and with 0.3\% and 76.6\% gain on DGCNN respectively. In the local optimization method, our L3A achieves the best performance on PointNet, PointNet++ and DGCNN. Moreover, Kim et.al. and Nudge are concurrent local attack work as our method, but our L3A is more cost-effective and achieves much better attack performance than them.

\section{Conclusion}
In this paper, we analyze existing weaknesses in global-optimization-based attack and further propose a novel local aggressive adversarial attack to solve them. To encourage the cost-effectiveness of points between attack and perceptibility, a salient subset of point cloud is selected to optimize toward the direction of preferred wrong class. Besides, to avoid falling in local optimality, PWA could adjust the attack and perceptibility purpose of each point by releasing reasonable budget from insignificant points. Extensive experiments demonstrate the state-of-the-art performance of our method on PointNet, PointNet++ and DGCNN.

\section*{Acknowledge}
This work was supported by the National Natural Science Foundation of China (Nos. 62076139, 61702280), the National Postdoctoral Program for Innovative Talents (No. BX20180146), China Postdoctoral Science Foundation (No. 2019M661901), Jiangsu Planned Projects for Postdoctoral Research Funds (No. 2019K024), Zhejiang Lab (No. 2021KF0AB05), 1311 Talent Program of Nanjing University of Posts and Telecommunications, and Natural Science Foundation of Jiangsu Province (No. BK20170900).

\bibliography{reference}

\begin{thebibliography}{42}
\providecommand{\natexlab}[1]{#1}
\providecommand{\url}[1]{\texttt{#1}}
\expandafter\ifx\csname urlstyle\endcsname\relax
  \providecommand{\doi}[1]{doi: #1}\else
  \providecommand{\doi}{doi: \begingroup \urlstyle{rm}\Url}\fi

\bibitem[Athalye et~al.(2018)Athalye, Engstrom, Ilyas, and Kwok]{ref27}
Anish Athalye, Logan Engstrom, Andrew Ilyas, and Kevin Kwok.
\newblock Synthesizing robust adversarial examples.
\newblock In \emph{ICML}, pages 284--293, 2018.

\bibitem[Beltr{\'a}n et~al.(2018)Beltr{\'a}n, Guindel, Moreno, Cruzado, Garcia,
  and De~La~Escalera]{ref38}
Jorge Beltr{\'a}n, Carlos Guindel, Francisco~Miguel Moreno, Daniel Cruzado,
  Fernando Garcia, and Arturo De~La~Escalera.
\newblock Birdnet: a 3d object detection framework from lidar information.
\newblock In \emph{ICITS}, pages 3517--3523, 2018.

\bibitem[Biggio et~al.(2013)Biggio, Corona, Maiorca, Nelson, {\v{S}}rndi{\'c},
  Laskov, Giacinto, and Roli]{ref8}
Battista Biggio, Igino Corona, Davide Maiorca, Blaine Nelson, Nedim
  {\v{S}}rndi{\'c}, Pavel Laskov, Giorgio Giacinto, and Fabio Roli.
\newblock Evasion attacks against machine learning at test time.
\newblock In \emph{Joint European Conference on Machine Learning and Knowledge
  Discovery in Databases}, pages 387--402, 2013.

\bibitem[Borgefors(1988)]{ref30}
Gunilla Borgefors.
\newblock Hierarchical chamfer matching: A parametric edge matching algorithm.
\newblock \emph{TPAMI}, 10\penalty0 (6):\penalty0 849--865, 1988.

\bibitem[Brown et~al.(2017)Brown, Man{\'e}, Roy, Abadi, and Gilmer]{ref14}
Tom~B Brown, Dandelion Man{\'e}, Aurko Roy, Mart{\'\i}n Abadi, and Justin
  Gilmer.
\newblock Adversarial patch.
\newblock \emph{arXiv preprint arXiv:1712.09665}, 2017.

\bibitem[Carlini and Wagner(2017)]{ref22}
Nicholas Carlini and David Wagner.
\newblock Towards evaluating the robustness of neural networks.
\newblock In \emph{IEEE Symposium on Security and Privacy}, pages 39--57, 2017.

\bibitem[Chang et~al.(2015)Chang, Funkhouser, Guibas, Hanrahan, Huang, Li,
  Savarese, Savva, Song, and Su]{ref13}
A.~X. Chang, T.~Funkhouser, L.~Guibas, P.~Hanrahan, Q.~Huang, Z.~Li,
  S.~Savarese, M.~Savva, S.~Song, and H.~Su.
\newblock Shapenet: An information-rich 3d model repository.
\newblock \emph{Computer Science}, 2015.

\bibitem[Dubuisson and Jain(1994)]{ref29}
M-P Dubuisson and Anil~K Jain.
\newblock A modified hausdorff distance for object matching.
\newblock In \emph{ICCV}, pages 566--568, 1994.

\bibitem[Geiger et~al.(2013)Geiger, Lenz, Stiller, and Urtasun]{ref15}
Andreas Geiger, Philip Lenz, Christoph Stiller, and Raquel Urtasun.
\newblock Vision meets robotics: The kitti dataset.
\newblock \emph{The International Journal of Robotics Research}, 32\penalty0
  (11):\penalty0 1231--1237, 2013.

\bibitem[Goodfellow et~al.(2014)Goodfellow, Shlens, and Szegedy]{ref9}
Ian~J Goodfellow, Jonathon Shlens, and Christian Szegedy.
\newblock Explaining and harnessing adversarial examples.
\newblock \emph{arXiv preprint arXiv:1412.6572}, 2014.

\bibitem[Kim et~al.(2020)Kim, Hua, Nguyen, and Yeung]{ref42}
Jaeyeon Kim, Binh-Son Hua, Duc~Thanh Nguyen, and Sai-Kit Yeung.
\newblock Minimal adversarial examples for deep learning on 3d point clouds.
\newblock \emph{arXiv preprint arXiv:2008.12066}, 2020.

\bibitem[Kurakin et~al.(2016{\natexlab{a}})Kurakin, Goodfellow, and
  Bengio]{ref11}
Alexey Kurakin, Ian Goodfellow, and Samy Bengio.
\newblock Adversarial machine learning at scale.
\newblock \emph{arXiv preprint arXiv:1611.01236}, 2016{\natexlab{a}}.

\bibitem[Kurakin et~al.(2016{\natexlab{b}})Kurakin, Goodfellow, Bengio,
  et~al.]{ref10}
Alexey Kurakin, Ian Goodfellow, Samy Bengio, et~al.
\newblock Adversarial examples in the physical world, 2016{\natexlab{b}}.

\bibitem[Li et~al.(2016)Li, Zhang, and Xia]{ref40}
Bo~Li, Tianlei Zhang, and Tian Xia.
\newblock Vehicle detection from 3d lidar using fully convolutional network.
\newblock \emph{arXiv preprint arXiv:1608.07916}, 2016.

\bibitem[Liang et~al.(2019)Liang, Yang, Chen, Hu, and Urtasun]{ref25}
Ming Liang, Bin Yang, Yun Chen, Rui Hu, and Raquel Urtasun.
\newblock Multi-task multi-sensor fusion for 3d object detection.
\newblock In \emph{CVPR}, pages 7345--7353, 2019.

\bibitem[Liu et~al.(2019)Liu, Yu, and Su]{ref17}
Daniel Liu, Ronald Yu, and Hao Su.
\newblock Extending adversarial attacks and defenses to deep 3d point cloud
  classifiers.
\newblock In \emph{ICIP}, pages 2279--2283, 2019.

\bibitem[Liu et~al.(2020)Liu, Yu, and Su]{ref23}
Daniel Liu, Ronald Yu, and Hao Su.
\newblock Adversarial shape perturbations on 3d point clouds.
\newblock In \emph{ECCV}, pages 88--104. Springer, 2020.

\bibitem[Liu et~al.(2018)Liu, Yang, Liu, Song, Li, and Chen]{ref32}
Xin Liu, Huanrui Yang, Ziwei Liu, Linghao Song, Hai Li, and Yiran Chen.
\newblock Dpatch: An adversarial patch attack on object detectors.
\newblock \emph{arXiv preprint arXiv:1806.02299}, 2018.

\bibitem[Madry et~al.(2017)Madry, Makelov, Schmidt, Tsipras, and Vladu]{ref12}
Aleksander Madry, Aleksandar Makelov, Ludwig Schmidt, Dimitris Tsipras, and
  Adrian Vladu.
\newblock Towards deep learning models resistant to adversarial attacks.
\newblock \emph{arXiv preprint arXiv:1706.06083}, 2017.

\bibitem[Mandikal et~al.(2018)Mandikal, Navaneet, Agarwal, and Babu]{ref41}
Priyanka Mandikal, KL~Navaneet, Mayank Agarwal, and R~Venkatesh Babu.
\newblock 3d-lmnet: Latent embedding matching for accurate and diverse 3d point
  cloud reconstruction from a single image.
\newblock \emph{arXiv preprint arXiv:1807.07796}, 2018.

\bibitem[Qi et~al.(2016)Qi, Su, Nie{\ss}ner, Dai, Yan, and Guibas]{ref2}
Charles~R Qi, Hao Su, Matthias Nie{\ss}ner, Angela Dai, Mengyuan Yan, and
  Leonidas~J Guibas.
\newblock Volumetric and multi-view cnns for object classification on 3d data.
\newblock In \emph{CVPR}, pages 5648--5656, 2016.

\bibitem[Qi et~al.(2017{\natexlab{a}})Qi, Su, Mo, and Guibas]{ref3}
Charles~R Qi, Hao Su, Kaichun Mo, and Leonidas~J Guibas.
\newblock Pointnet: Deep learning on point sets for 3d classification and
  segmentation.
\newblock In \emph{CVPR}, pages 652--660, 2017{\natexlab{a}}.

\bibitem[Qi et~al.(2017{\natexlab{b}})Qi, Yi, Su, and Guibas]{ref4}
Charles~R Qi, Li~Yi, Hao Su, and Leonidas~J Guibas.
\newblock Pointnet++: Deep hierarchical feature learning on point sets in a
  metric space.
\newblock \emph{arXiv preprint arXiv:1706.02413}, 2017{\natexlab{b}}.

\bibitem[Rusu et~al.(2008)Rusu, Marton, Blodow, Dolha, and Beetz]{ref16}
Radu~Bogdan Rusu, Zoltan~Csaba Marton, Nico Blodow, Mihai Dolha, and Michael
  Beetz.
\newblock Towards 3d point cloud based object maps for household environments.
\newblock \emph{Robotics and Autonomous Systems}, 56\penalty0 (11):\penalty0
  927--941, 2008.

\bibitem[Shi et~al.(2019)Shi, Wang, and Li]{ref26}
Shaoshuai Shi, Xiaogang Wang, and Hongsheng Li.
\newblock Pointrcnn: 3d object proposal generation and detection from point
  cloud.
\newblock In \emph{CVPR}, pages 770--779, 2019.

\bibitem[Su et~al.(2015)Su, Maji, Kalogerakis, and Learned-Miller]{ref1}
Hang Su, Subhransu Maji, Evangelos Kalogerakis, and Erik Learned-Miller.
\newblock Multi-view convolutional neural networks for 3d shape recognition.
\newblock In \emph{ICCV}, pages 945--953, 2015.

\bibitem[Szegedy et~al.(2013)Szegedy, Zaremba, Sutskever, Bruna, Erhan,
  Goodfellow, and Fergus]{ref7}
Christian Szegedy, Wojciech Zaremba, Ilya Sutskever, Joan Bruna, Dumitru Erhan,
  Ian Goodfellow, and Rob Fergus.
\newblock Intriguing properties of neural networks.
\newblock \emph{arXiv preprint arXiv:1312.6199}, 2013.

\bibitem[Thys et~al.(2019)Thys, Van~Ranst, and Goedem{\'e}]{ref31}
Simen Thys, Wiebe Van~Ranst, and Toon Goedem{\'e}.
\newblock Fooling automated surveillance cameras: adversarial patches to attack
  person detection.
\newblock In \emph{CVPRW}, pages 1--6, 2019.

\bibitem[Wang et~al.(2019)Wang, Sun, Liu, Sarma, Bronstein, and Solomon]{ref5}
Yue Wang, Yongbin Sun, Ziwei Liu, Sanjay~E Sarma, Michael~M Bronstein, and
  Justin~M Solomon.
\newblock Dynamic graph cnn for learning on point clouds.
\newblock \emph{ACM TOG}, 38\penalty0 (5):\penalty0 1--12, 2019.

\bibitem[Wicker and Kwiatkowska(2019)]{ref33}
Matthew Wicker and Marta Kwiatkowska.
\newblock Robustness of 3d deep learning in an adversarial setting.
\newblock In \emph{CVPR}, pages 11767--11775, 2019.

\bibitem[Wu et~al.(2016)Wu, Zhang, Xue, Freeman, and Tenenbaum]{ref37}
Jiajun Wu, Chengkai Zhang, Tianfan Xue, William~T Freeman, and Joshua~B
  Tenenbaum.
\newblock Learning a probabilistic latent space of object shapes via 3d
  generative-adversarial modeling.
\newblock \emph{arXiv preprint arXiv:1610.07584}, 2016.

\bibitem[Wu et~al.(2015{\natexlab{a}})Wu, Song, Khosla, Yu, Zhang, Tang, and
  Xiao]{modelnet40}
Zhirong Wu, Shuran Song, Aditya Khosla, Fisher Yu, Linguang Zhang, Xiaoou Tang,
  and Jianxiong Xiao.
\newblock 3d shapenets: A deep representation for volumetric shapes.
\newblock In \emph{CVPR}, pages 1912--1920, 2015{\natexlab{a}}.

\bibitem[Wu et~al.(2015{\natexlab{b}})Wu, Song, Khosla, Yu, Zhang, Tang, and
  Xiao]{ref6}
Zhirong Wu, Shuran Song, Aditya Khosla, Fisher Yu, Linguang Zhang, Xiaoou Tang,
  and Jianxiong Xiao.
\newblock 3d shapenets: A deep representation for volumetric shapes.
\newblock In \emph{CVPR}, pages 1912--1920, 2015{\natexlab{b}}.

\bibitem[Wu et~al.(2020)Wu, Duan, Wang, Fan, and Guibas]{ref24}
Ziyi Wu, Yueqi Duan, He~Wang, Qingnan Fan, and Leonidas~J Guibas.
\newblock If-defense: 3d adversarial point cloud defense via implicit function
  based restoration.
\newblock \emph{arXiv preprint arXiv:2010.05272}, 2020.

\bibitem[Xiang et~al.(2019{\natexlab{a}})Xiang, Qi, and Li]{ref18}
Chong Xiang, Charles~R Qi, and Bo~Li.
\newblock Generating 3d adversarial point clouds.
\newblock In \emph{CVPR}, pages 9136--9144, 2019{\natexlab{a}}.

\bibitem[Xiang et~al.(2019{\natexlab{b}})Xiang, Qi, and Li]{ref34}
Chong Xiang, Charles~R Qi, and Bo~Li.
\newblock Generating 3d adversarial point clouds.
\newblock In \emph{CVPR}, pages 9136--9144, 2019{\natexlab{b}}.

\bibitem[Yang et~al.(2018)Yang, Luo, and Urtasun]{ref39}
Bin Yang, Wenjie Luo, and Raquel Urtasun.
\newblock Pixor: Real-time 3d object detection from point clouds.
\newblock In \emph{CVPR}, pages 7652--7660, 2018.

\bibitem[Zhang et~al.(2019)Zhang, Yang, Fang, Ni, Liu, and Tian]{ref19}
Qiang Zhang, Jiancheng Yang, Rongyao Fang, Bingbing Ni, Jinxian Liu, and
  Qi~Tian.
\newblock Adversarial attack and defense on point sets.
\newblock \emph{arXiv preprint arXiv:1902.10899}, 2019.

\bibitem[Zhao et~al.(2020)Zhao, Shumailov, Mullins, and Anderson]{ref43}
Yiren Zhao, Ilia Shumailov, Robert Mullins, and Ross Anderson.
\newblock Nudge attacks on point-cloud dnns.
\newblock \emph{arXiv preprint arXiv:2011.11637}, 2020.

\bibitem[Zheng et~al.(2019)Zheng, Chen, Yuan, Li, and Ren]{ref20}
Tianhang Zheng, Changyou Chen, Junsong Yuan, Bo~Li, and Kui Ren.
\newblock Pointcloud saliency maps.
\newblock In \emph{ICCV}, pages 1598--1606, 2019.

\bibitem[Zhou et~al.(2019)Zhou, Chen, Zhang, Fang, Zhou, and Yu]{ref21}
Hang Zhou, Kejiang Chen, Weiming Zhang, Han Fang, Wenbo Zhou, and Nenghai Yu.
\newblock Dup-net: Denoiser and upsampler network for 3d adversarial point
  clouds defense.
\newblock In \emph{ICCV}, pages 1961--1970, 2019.

\bibitem[Zhou and Tuzel(2018)]{ref28}
Yin Zhou and Oncel Tuzel.
\newblock Voxelnet: End-to-end learning for point cloud based 3d object
  detection.
\newblock In \emph{CVPR}, pages 4490--4499, 2018.

\end{thebibliography}

\end{document}